\documentclass[submission,copyright]{eptcs}

\newtheorem{definition}{Definition}

\newcommand{\textcite}[1]{\citeauthor{#1}'s \cite{#1}}

\usepackage[utf8]{inputenc}
\usepackage[T1]{fontenc}
\usepackage{amsmath}
\usepackage[sc]{mathpazo}
\usepackage[american]{babel}
\usepackage{graphicx}
\usepackage{listings}
\usepackage{lstautogobble}
\usepackage{tikz}
\usepackage{pgfplots}
\usetikzlibrary{patterns}
\usepackage{pgfplotstable}
\usepackage[font={footnotesize}]{caption}
\usepackage{enumitem}
\usepackage{color}
\usepackage{todonotes}
\usepackage{multirow}
\usepackage{longtable}
\usepackage{subfig}
\usepackage{wrapfig}

\pgfplotsset{compat=1.13} 
\pgfplotsset{
  cycle list={CornflowerBlue\\Dandelion\\ForestGreen\\BrickRed\\},
}
\lstdefinelanguage{XES-XML}{
  language     = xml,
  morekeywords = {event, string, date, log, extension, trace},
}

\colorlet{punct}{red!60!black}
\definecolor{delim}{RGB}{20,105,176}
\colorlet{numb}{magenta!60!black}
\lstdefinelanguage{json}{
    basicstyle=\normalfont\ttfamily,
    numbers=left,
    numberstyle=\scriptsize,
    comment=[l]{//},
    morecomment=[s]{/*}{*/},
    commentstyle=\color{purple}\ttfamily,
    stepnumber=1,
    numbersep=8pt,
    showstringspaces=false,
    breaklines=true,
    frame=lines,
    literate=
     *{0}{{{\color{numb}0}}}{1}
      {1}{{{\color{numb}1}}}{1}
      {2}{{{\color{numb}2}}}{1}
      {3}{{{\color{numb}3}}}{1}
      {4}{{{\color{numb}4}}}{1}
      {5}{{{\color{numb}5}}}{1}
      {6}{{{\color{numb}6}}}{1}
      {7}{{{\color{numb}7}}}{1}
      {8}{{{\color{numb}8}}}{1}
      {9}{{{\color{numb}9}}}{1}
      {:}{{{\color{punct}{:}}}}{1}
      {,}{{{\color{punct}{,}}}}{1}
      {\{}{{{\color{delim}{\{}}}}{1}
      {\}}{{{\color{delim}{\}}}}}{1}
      {[}{{{\color{delim}{[}}}}{1}
      {]}{{{\color{delim}{]}}}}{1},
}

\lstdefinestyle{terminal}
{
    basicstyle=\small\ttfamily,
    frame=none,
    numbers=none,
}

\definecolor{pblue}{rgb}{0.13,0.13,1}
\definecolor{pgreen}{rgb}{0,0.5,0}
\definecolor{pred}{rgb}{0.9,0,0}
\definecolor{pgrey}{rgb}{0.46,0.45,0.48}

\lstdefinestyle{java}
{
  language=Java,
  numbers=left,
  breaklines=true,
  frame=lines,
  stepnumber=1,
  numbersep=8pt,
  numberstyle=\scriptsize,
  showspaces=false,
  showtabs=false,
  showstringspaces=false,
  breakatwhitespace=true,
  commentstyle=\color{pgreen},
  keywordstyle=\color{pblue},
  stringstyle=\color{pred},
  basicstyle=\ttfamily,
  moredelim=[il][\textcolor{pgrey}]{$$},
  moredelim=[is][\textcolor{pgrey}]{\%\%}{\%\%}
}

\pgfplotsset{
  counter_barchart/.style={
    width=0.6\textwidth,
    height=5cm,
    xbar,
    xmin=0,
    xmajorgrids = true,
    tick align = outside, xtick pos = left,
    x tick label style={/pgf/number format/1000 sep=},
    scaled x ticks=false,
    enlarge y limits=0.4,
    enlarge x limits={0.15,upper},
    symbolic y coords={modified, classic},
    ytick=data,
    yticklabel style={text width=0.125\textwidth,align=flush left},
    nodes near coords,
    nodes near coords align={horizontal},
    legend style={at={(1.05,0.24)},anchor=south west},
    reverse legend
  }
}

\definecolor{newRed}{RGB}{231, 76, 60}
\definecolor{newGreen}{RGB}{39, 174, 96}
\definecolor{newBlue}{RGB}{41, 128, 185}

\pgfplotscreateplotcyclelist{mycolorlist}{
{newRed, fill=newRed},
{newGreen, fill=newGreen},
{newBlue, fill=newBlue},
{newRed, pattern color = newRed, pattern = north west lines},
{newGreen, pattern color = newGreen, pattern = north west lines},
{newBlue, pattern color = newBlue, pattern = north west lines},
}

\pgfplotsset{
  counter_barchart_y/.style={
    bar width = 2mm,
    width=0.45\textwidth,
    height=3.6cm,
    ybar,
    ymin=0,
    ymax = 1,
    ymajorgrids = true,
    tick align = outside, ytick pos = left,
    y tick label style={/pgf/number format/1000 sep=, font=\tiny},
    scaled y ticks=false,
    enlarge x limits=0.27,
    enlarge y limits={0.49,upper},
    symbolic x coords={Gößler and Métayer, Gößler and Astefanoaei, Policy Compliance Algorithm},
    xtick=data,
    xticklabel style={text width=0.13\textwidth,align=center, font=\tiny},
    nodes near coords,
    nodes near coords align={horizontal},
    every node near coord/.append style={
                font=\tiny,
                anchor=west,
                rotate=90,
                /pgf/number format/fixed zerofill,
    			 /pgf/number format/precision=3
        },
    legend style={at={(0.5,-0.38)},anchor=north, legend columns = -1, font=\tiny},
    cycle list name=mycolorlist
  }
}

\newcolumntype{C}[1]{>{\centering\arraybackslash}p{#1}}

\providecommand{\event}{CREST 2017} 

\title{ACCBench: A Framework for Comparing Causality Algorithms}
\author{Simon Rehwald \quad Amjad Ibrahim \quad Kristian Beckers \quad  Alexander Pretschner
\institute{Department of Informatics, Technical University of Munich, Garching b. Munich, Germany}
\email{\{rehwald, ibrahim, beckers, pretschn\}@in.tum.de}
}
\def\titlerunning{ACCBench: A Framework for Comparing Causality Algorithms}
\def\authorrunning{S. Rehwald, A. Ibrahim, K. Beckers, A. Pretschner }
\def\copyrightholders{S. Rehwald, A. Ibrahim,\\ K. Beckers, A. Pretschner }
\begin{document}
\maketitle

\begin{abstract}
Modern socio-technical systems are increasingly complex. A fundamental problem is that the borders of such systems are often not well-defined a-priori, which among other problems can lead to unwanted behavior during runtime. Ideally, unwanted behavior should  be prevented. If this is not possible the system shall at least be able to help determine potential cause(s) a-posterori, identify responsible parties and make them accountable for their behavior. Recently, several algorithms addressing these concepts have been proposed. However, the applicability of the corresponding approaches, specifically their effectiveness and performance, is mostly unknown. Therefore, in this paper, we propose \textit{ACCBench}, a benchmark tool that allows to compare and evaluate causality algorithms under a consistent setting. Furthermore, we contribute an implementation of the two causality algorithms by \cite{goessler2013generaltrace} and \cite{goessler2014blaming} as well as of a policy compliance approach based on some concepts of \cite{mian2015auditing}. Lastly, we conduct a case study of an Intelligent Door Control System, which exposes concrete strengths and weaknesses of all algorithms under different aspects. In the course of this, we show that the effectiveness of the algorithms in terms of cause detection as well as their performance differ to some extent. In addition, our analysis reports on some qualitative aspects that should be considered when evaluating each algorithm. For example, the human effort needed to configure the algorithm and model the use case is analyzed.
\end{abstract}

\section{Introduction}\label{section:introduction}
Our society, industry and daily lives are built on increasingly complex systems, be it artificial organs or autonomous vehicles. Most of these consist of multiple or even hundreds of components, all interacting with each other. These systems are increasingly vulnerable due to, among others, cyber attacks, bugs, defective hardware, which in turn pose potential risk to the economy as well as people’s health and livelihood. We must prevent these failures before their occurrence; at the very least, we must be able to determine \textit{why} a system failed and \textit{what}/\textit{who} caused the failure.

\begin{wrapfigure}{R}{0.4\textwidth}
\vspace{-3mm}
\centering
\includegraphics[trim=6mm 6mm 6mm 6mm, clip, width=0.4\textwidth, keepaspectratio]{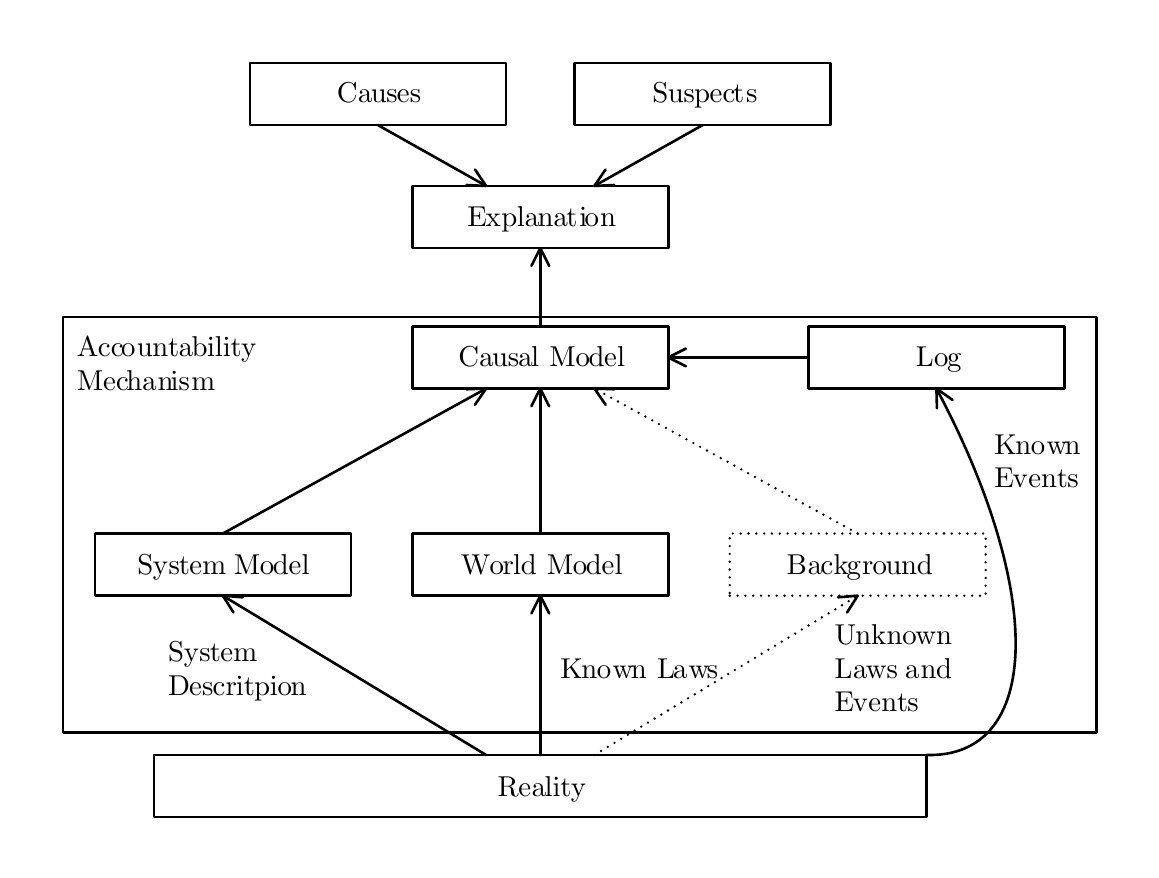}
\caption[Overview of Accountability (adopted from \cite{kacianka2016unified})]{Overview of Accountability (adopted from \cite{kacianka2016unified}; arrows in this model need to be read as ``is represented in'')}
\label{fig:accountability_overiview}
\vspace{-4mm}
\end{wrapfigure}

A term often used in the context of these problems, is \textit{accountability} \cite{Beckers2016, weitzner2008information}. In computer science, accountability is usually seen as a property of systems, which allows one to link actions and activities to specific parts and actors of a system and hold the latter liable for potential misbehavior. From a legal perspective, accountability is becoming increasingly important as for instance evidenced by autonomous driving systems. The behavior of such systems is unpredictable during design time due to their self-learning nature. Hence, we think that accountability is a property which will improve our current technological abilities. One way how systems can enable accountability is the usage of event logs. Yet, because of their complexity, manually analyzing these logs is practically impossible. Thus, reasonable mechanisms, which automatically determine responsible parties for specific observations or at least support humans during this analysis, are needed.

\noindent
Recently, \cite{kacianka2016unified} have proposed a high level overview of accountability. As we can see in Fig. \ref{fig:accountability_overiview}, also \textit{causality} \cite{halpern2005causes_1, halpern2005causes_2, halpern2015modification} plays an important role in an accountability mechanism. Using knowledge about the environment and the system itself as well as the current events, a causal model is created. This model allows to identify explanation(s) for the observed behavior and misbehavior of a system.
\par
In current research, several algorithms claiming to be capable of generating those causal models have been proposed and evaluated in different case studies. However, we neither know the quality of these approaches in the sense of performance and effectiveness, nor do we know, whether and to what extent they are applicable. This paper therefore makes the following contributions:
\vspace{-0.5mm}
\begin{itemize}
\itemsep-0em 
\renewcommand\labelitemi{--}
\item  An extensible benchmark tool called \textit{ACCBench} (\textit{Accountability Causality Comparison Benchmark}) for comparing and evaluating causality algorithms regarding the criteria performance and effectiveness is developed. We propose metrics base on binary classification for estimating the latter criterion.
\item The implementation of two causality algorithms (based on \cite{goessler2013generaltrace} and \cite{goessler2014blaming}) and one policy compliance algorithm (based on \cite{mian2015auditing}), all of which perform their analysis using event logs.
\item The algorithms are compared and evaluated in a case study. It considers Door Control System. The event logs for this system have been generated with \textit{CPNTools}\footnote{\url{http://cpntools.org/} [Accessed 07 February 2017]} using an approach similar to \cite{medeiros2005process}.
\end{itemize}
\vspace{-0.5mm}
Our choice of the three aforementioned algorithms is due in part to the fact that they are among the most current in today’s research. In addition, their concepts seem more implementable thanks to their technical nature. Moreover, comparing a state-of-the-art causality algorithm with a policy compliance algorithm may point out interesting differences between both approaches.
\par
The remainder of this paper is structured as follows. We start with related work in Section \ref{section:related_work}. Subsequently, we describe the new benchmark tool \textit{ACCBench}, explain our used metrics and briefly consider the implemented algorithms (Section \ref{section:accbench}). Then, in Section \ref{section:comparison}, we present our case study containing the analysis and comparison of the implemented algorithms. Finally, we conclude this paper in Section \ref{section:conclusion}.
\section{Related Work}\label{section:related_work}
The term \textit{accountability} has been described in multiple different ways (see for instance \cite{mulgan2000accountability, bovens2010two,weitzner2008information}). In this paper however, we refer to the definition used in \cite{Beckers2016}. The authors think of \textit{accountability} as a capability of a socio-technical systems to answer questions regarding the cause of occurred unwanted behavior. 
Similarly, \textit{causality} is a term with a variety of definitions. One possible understanding of causality is the counterfactual approach, which tries to establish causal relationships by asking questions like ``Would have event $B$ still happened if another event $A$ would have not happened?''. A widely-adopted formal definition based on the counterfactual approach has been made in \cite{halpern2005causes_1, halpern2005causes_2, halpern2015modification}.

To the best of our knowledge, the work about the comparison of causality algorithms is insufficient. 
One related work compares the approaches to determine causality between variables. In \cite{mooij2016distinguishing}, the authors analyze a set of bivariate causal discovery methods. These methods compute which variable out of two is the cause for the other variable, i.e. the causal relationship between two variables is searched. The methods are compared by checking whether or not they can detect the correct causing variable using the available data. The key difference of this work to our approach is that we do not compare techniques for discovering the causal relationship between two variables, but techniques for discovering the cause(s) for observed faulty behavior of a system. Nevertheless, the main idea, i.e. the comparison of different causal methodologies and their evaluation, stays the same. Besides that, \cite{guyon2009causality, guyon2011causality, guyon2011time} describe the ``Causality Workbench'' project, which provides resources for benchmarking causal discovery algorithms. These resources comprise sample datasets and software tools. Furthermore a virtual  lab \cite{guyon2009causality} has been created, which allows to conduct experiments in order to analyze the causal relationships in a provided model. Similar to the approach in  \cite{mooij2016distinguishing}, that project differs from the goals of this paper in several parts as well. First of all, the ``Causality Workbench'' just provides the sample data for comparing different algorithms. However, the actual comparison and the development of the criteria need to be performed by the user. In \textit{ACCBench}, in contrast, the comparison criteria are built into the software, but the user needs to provide the data. Moreover, the ``Causality Workbench'' is again designed for causal discovery algorithms.
\section{ACCBench: Accountability Causality Comparison Benchmark}\label{section:accbench}
In this section, we describe the benchmark tool: \textit{ACCBench}. We give an overview regarding the architecture and general structure  of \textit{ACCBench} (Section \ref{subsection:overview}). Subsequently, we present the quantitative metrics used for comparing the causality algorithms (Section \ref{subsection:metrics}). Lastly, in Section \ref{subsection:algorithms}, we have a closer look at the algorithms implemented as part of \textit{ACCBench} and the adaptations we made.

\subsection{Overview}\label{subsection:overview}
As we mentioned before, accountability is desirable, if a (part of a) system did not behave as expected. That is, behavior, which does not correspond to the``normal'' behavior of that (part of a) system. We summarize such a situation with \textit{unwanted behavior}:
\begin{definition}\label{def:unwanted_behavior}
\textnormal{\textbf{(Unwanted behavior).}} Unwanted behavior represents a deviation of a system's normal and intended behavior. Unwanted behavior may cause security, safety and/or privacy issues within a system.
\end{definition}
For example, unwanted behavior in an airplane would be that the landing gear is not extended although the respective commands were sent to the control unit. Taking a look at the recorded behavior in the form of event logs, reasonable mechanism should be capable of finding the cause(s) for the unwanted behavior.

With the help of \textit{ACCBench}, we want to compare multiple causality mechanisms (algorithms) regarding their performance and effectiveness in finding the cause of an unwanted behavior (Def. \ref{def:unwanted_behavior}). Therefore, we need to (1) create a consistent setting, i.e. an environment, in which all algorithms have the same assumptions and information about a system, as well as (2) reasonable metrics.
\begin{wrapfigure}{R}{0.45\textwidth}
\vspace{-3mm}
\centering
\includegraphics[trim=6mm 6mm 6mm 6mm, clip, width=0.45\textwidth, keepaspectratio]{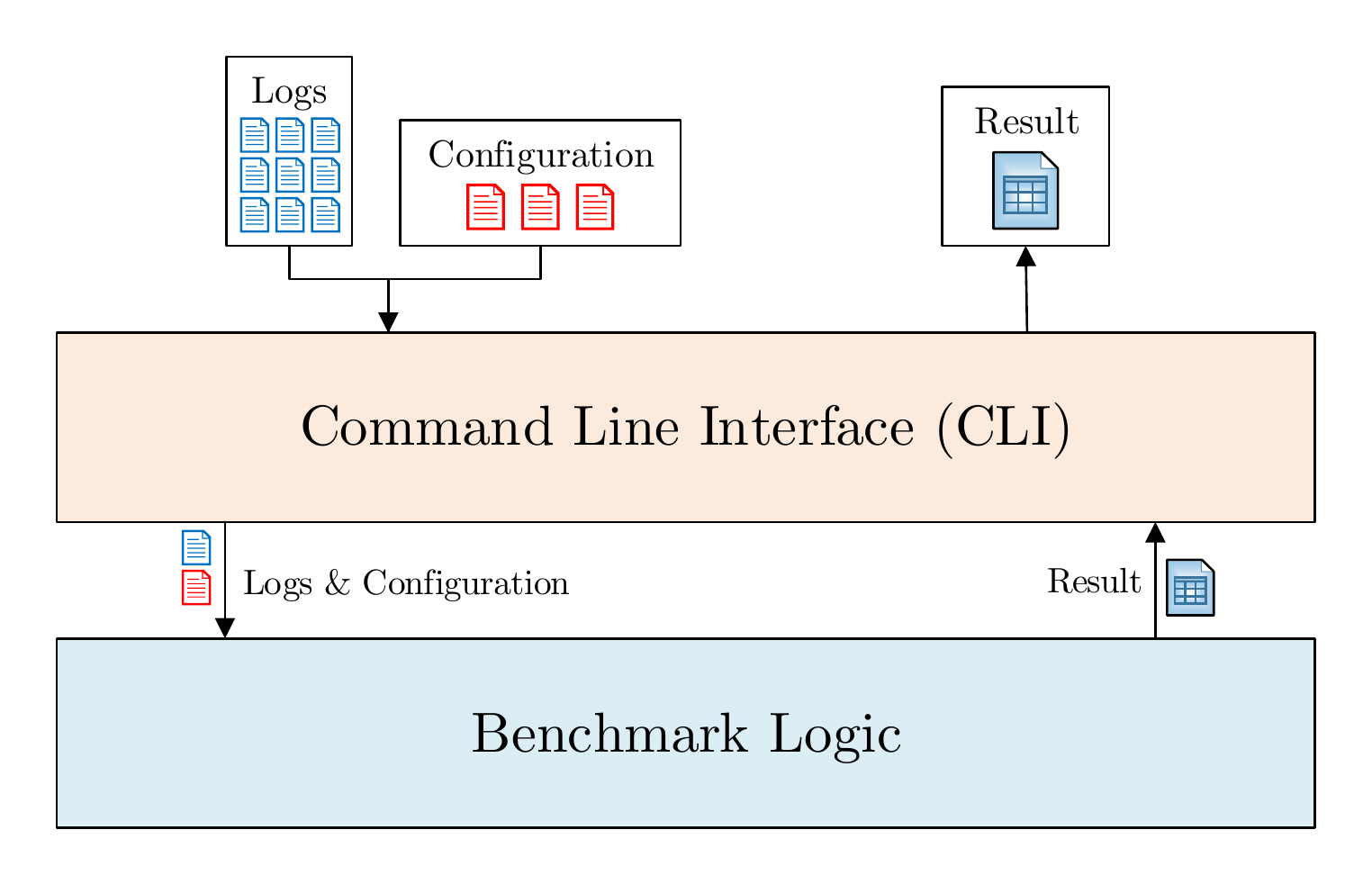}
\caption[Informal Model of ACCBench]{Informal Model of \textit{ACCBench}}
\label{fig:accbench:informal_model}
\vspace{-4mm}
\end{wrapfigure}
In Fig. \ref{fig:accbench:informal_model}, the basic idea and functionality of this tool are illustrated.

\noindent
On the one hand, a user needs to provide a set of log files captured by the system, in which the cause(s) for potentially observed unwanted behavior shall be identified. On the other hand, a file containing information an algorithm uses to understand the behavior of the system, i.e. the system/world model (cf. Fig. \ref{fig:accountability_overiview}) is necessary. We refer to such a file as \textit{configuration} file. The specified event logs and configuration files are then passed to the Benchmark Logic layer, which executes the algorithms. Finally, the benchmark result is presented to the user.

Technically, \textit{ACCBench} (effectiveness metrics and algorithms ) is implemented in \textit{Java}. Also, \textit{Java Microbenchmark Harness (JMH)}\footnote{\url{http://openjdk.java.net/projects/code-tools/jmh/} [Accessed 07 February 2017]} is used  to profile the algorithms and measure the performance. This microbenchmarking framework allows to cope with common pitfalls when making measurements on the JVM, e.g. optimization, garbage collection etc., and leads to more accurate results. Besides that, we also took advantage of an existing format for event logs and a corresponding parser, namely an XML-based format called \textit{Extensible Event Stream (XES)}\footnote{\url{http://www.xes-standard.org} [Accessed 07 February 2017]}.

\subsection{Metrics}\label{subsection:metrics}
In this subsection, we propose our criteria for measuring the effectiveness of an algorithm, i.e. the degree to which it returns "correct" results. Next we discuss measuring the performance.

\subsubsection{Effectiveness}
Before reasoning about the correctness of the result of a causality algorithm, we need to define such a result. Similarly as \cite{halpern2005causes_1, halpern2005causes_2, halpern2015modification}, in this paper we consider a \textit{cause} as a conjunction of events. Thus, only the occurrence of all these events causes the observed unwanted behavior (Def. \ref{def:unwanted_behavior}). For convenience, we model this conjunction as a set of events. Moreover, there might be multiple causes for unwanted behavior. Therefore, we assume that a causality algorithm returns a set of sets of events, where each set represents one unique cause. We aim to compare the correct/expected causes ($\mathcal{R}^{\text{exp}}$) with the ones an algorithm returns ($\mathcal{R}^{\text{hyp}}$). We have formalized $\mathcal{R}^{\text{exp}}$ and $\mathcal{R}^{\text{hyp}}$ in the following definition:
\begin{definition}\label{def:root_cause_mathematical}
\textnormal{\textbf{Cause $R$, expected (hypothesized) causes $\mathcal{R}^{\text{exp}}$ ($\mathcal{R}^{\text{hyp}}$)}} A cause $R = \{e_1, ..., e_n\} \subseteq L$ ($n \in \mathbb{N}$) is a set of events $e_i \in L$, where $L$ is the set of uniquely identifiable events in an event log. An expected set of causes $\mathcal{R}^{\text{exp}} = \{R_1, ..., R_j\}$ ($j \in \mathbb{N}$) is a set of causes $R_1, ..., R_j$, which represents the real causes for observed unwanted behavior. A hypothesized set of causes $\mathcal{R}^{\text{hyp}}= \{R_1, ..., R_k\}$ ($k \in \mathbb{N}$) is a set of causes $R_1, ..., R_k$, which represent the hypothesized causes for observed unwanted behavior.
\end{definition}
We are aware of the fact that this representation of causes does not take into account the specific order of events. However, in some cases not only the existence of events, but also their order defines whether or not they need to be considered as a cause. Therefore our approach should rather be considered as an approximation of the effectiveness. Notice furthermore that the specification of the real causes $\mathcal{R}^{\text{exp}}$ is highly dependent on the underlying definition of a cause. In our case study (Section \ref{section:comparison}) we derived those causes manually looking at the results of our threat analysis and relying on the definition of a cause made by the causality algorithms described in Section \ref{subsection:algorithms}.
\par
We found out that comparing the two sets $\mathcal{R}^{\text{exp}}$ and $\mathcal{R}^{\text{hyp}}$ is not necessarily trivial. The main problem to overcome is that we want to take partial correctness of results into account. The rationale behind this is that intuitively an algorithm whose reported causes do not fully, but partially, match the correct ones should be considered as more effective than an algorithm whose results are completely wrong. Nevertheless, it might sometimes also be desirable to only count fully correct results when benchmarking algorithms. Thus, we came up with two different solutions building upon each other.

\paragraph{``Normal'' Binary Classification}
This approach does not take partial correctness into account. Thus, we use a classification: true positives, false positives, true negatives and false negatives (Tab. \ref{tab:confusion_matrix}) in order to compute common metrics as precision, recall and $F_1$-measure (cf. \cite{facwett2006introduction, chinchor1992muc4}). Precision is defined as the amount of true positives divided by the number of predicted positives (i.e. the sum of true and false positives), whereas for computing the recall the amount of true positives is divided by the number of real positives (i.e. the sum of true positives and false negatives). 
\vspace*{-3mm}
\begin{table}[ht!]
\footnotesize
\hbadness=10000
\centering
\begin{tabular}{c c p{5cm} | p{5cm}}
&&\multicolumn{2}{c}{\textbf{Real Class}}\\ \cline{3-4}
&&\multicolumn{1}{|c|}{+R ($= \mathcal{R}^{\text{exp}}$)} &\multicolumn{1}{c|}{-R}\\ \cline{2-4}
\parbox[t]{3mm}{\multirow{6}{*}{\rotatebox[origin=c]{90}{\textbf{Predicted Class}}}} &\multicolumn{1}{|c|}{\multirow{4}{*}{\parbox[t]{1.25cm}{\centering +P ($= \mathcal{R}^{\text{hyp}}$)}}}
&\underline{\smash{True Positives}}: The causes reported by a causality algorithm, which are also causes in reality.
&\multicolumn{1}{p{5cm}|}{\underline{\smash{False Positives}}: The causes reported by a causality algorithm, which are \emph{not} causes in reality.} \\ \cline{2-4}
&\multicolumn{1}{|c|}{\multirow{4}{*}{\parbox[t]{1.25cm}{\centering -P}}} &\underline{\smash{False Negatives}}: The causes \emph{not} reported by a causality algorithm, which are actually causes in reality.
&\multicolumn{1}{p{5cm}|}{\underline{\smash{True Negatives}}: The causes \emph{not} reported by a causality algorithm, which are also \emph{not} causes in reality.}\\  \cline{2-4}
\end{tabular}
\caption{Binary Classification in the Context of Causality Algorithms}
\label{tab:confusion_matrix}
\end{table}
\vspace*{-3mm}
Hence, precision describes how many of the predicted positives are actually true positives and recall describes how many of the real positives have been correctly identified as positives. The $F_1$-measure \cite{chinchor1992muc4} is a metric combining both recall and precision:
\[F_1 = \frac{2\cdot\text{recall}\cdot\text{precision}}{\text{recall}+\text{precision}}\]
Using Tab. \ref{tab:confusion_matrix}, we classify each $R_i \in \mathcal{R}^{\text{hyp}}$ returned by an algorithm using the real positives $\mathcal{R}^{\text{exp}}$. Since we do not care about partial correctness, a returned cause $R_i \in \mathcal{R}^{hyp}$ is only then classified as true positive, if $R_i \in \mathcal{R}^{\text{exp}}$ holds, i.e. the complete set $R_i$ needs to be an element of the set of sets $\mathcal{R}^{\text{exp}}$. However, this is a very strict classification.

\paragraph{``Modified'' Binary Classification}
We developed a second approach for measuring the effectiveness. We reuse the idea of the \textit{Best Match} algorithm in \cite{goldberg2010measuring} and create a modified binary classification similar to Tab. \ref{tab:confusion_matrix}. Specifically, we relax the definitions of true positives etc. to allow real values instead of natural ones. For example, a ``0.5-true-positive'' reflects partially correct reported causes. The benefit is that all the previously mentioned metrics based on binary classification, e.g. precision, can be applied. Firstly, we compute a potentially real-valued number of true positives $N_{TP'}$, which we refer to as ``modified true positives'', and then use some relations in order to obtain the number of ``modified false positives'' $N_{FP'}$ and ``modified false negatives'' $N_{FN'}$. Notice that we compute \emph{numbers} and not the actual \emph{sets}\footnote{This would not even be possible, because we would possibly need to express partial membership of some elements to a set.}.
\par
Intuitively, we want to find the best match in the sense of highest similarity of each $R_i \in \mathcal{R}^{\text{hyp}}$ in $\mathcal{R}^{\text{exp}}$, yet under the condition that there are not any $R_i, R_j \in \mathcal{R}^{\text{hyp}}$, which are matched with the same $R_k \in \mathcal{R}^{\text{exp}}$. Put another way, we match at most \emph{one} $R_i \in \mathcal{R}^{\text{hyp}}$ to any $R_k \in \mathcal{R}^{\text{exp}}$. We can then sum up the (normalized) similarity between the assignments $(R_i, R_k)$ and obtain a real-valued number of true positives $N_{TP'}$. For the similarity $s(S, S')$ between two sets, we can reuse the formula in \cite{goldberg2010measuring} for the distance $d(S, S')$ between two sets and transform it to yield the similarity (or the Jaccard index, respectively):
\[
s(S, S') = 1 - d(S,S') = 1 - \left(1 - \frac{\vert S \cap S' \vert}{\vert S \cup S' \vert}\right) = \frac{\vert S \cap S' \vert}{\vert S \cup S' \vert}
\]
However, assigning $R_i \in \mathcal{R}^{\text{hyp}}$ to a $R_k \in \mathcal{R}^{\text{exp}}$ is not trivial; an assignment problem needs to be solved. Having computed $N_{TP'}$, we can use relations between +R, +P, TP, FP and FN, in order to derive the number of modified false positives $N_{FP'}$ and false negatives $N_{FN'}$:
\begin{itemize}
\item $N_{FN'} = \vert \mathcal{R}^{\text{exp}} \vert - N_{TP'}$
\item $N_{FP'} = \vert \mathcal{R}^{\text{hyp}} \vert - N_{TP'}$
\end{itemize}
As shown in Tab. \ref{tab:confusion_matrix}, the number of true positives +R, which is equal to $\mathcal{R}^{\text{exp}}$, can be obtained by calculating the corresponding column sum, i.e. the sum of true positives and false negatives. Similarly, we obtain the number of predicted positives +P, which is equal to $\mathcal{R}^{\text{hyp}}$, by the corresponding row sum, i.e. the sum of true positives and false positives. These relations remain unchanged, even if we allow floats for the true positives etc. Subsequently, we can insert the computed values for $N_{TP'}$, $N_{FP'}$ and $N_{FN'}$ in the formulas for precision, recall and $F_1$-measure without having to change anything in the latter. 

\subsubsection{Performance}
For measuring the performance of the causality algorithms, we use the execution time and amount of allocated memory for the analysis of a single event log. As mentioned before, \textit{ACCBench} makes sure that values for both are obtained in a reasonable and meaningful way by using \textit{JMH}.

\subsection{Algorithms}\label{subsection:algorithms}
In this section, we will briefly describe the concepts of the algorithms implemented in \textit{ACCBench}. For more detailed considerations the respective papers of these approaches should be consulted.

\paragraph{Gössler and Métayer \cite{goessler2013generaltrace}}
 In order to analyze causality within a system, the authors use a ``language-based modeling framework'' to model a system. That is, the behavior of each component and the system itself is defined by a formal language. Intuitively, a single character in the ``alphabet'' of a component represents an atomic event and a ``word'' describes a specific sequence of events. On that way, incorrect behavior of one or more components can be detected by checking whether or not the behavior of the latter is a valid word in the current language. The algorithm then replaces such incorrect behavior with correct behavior taken from the specification given as formal language. If the unwanted behavior cannot be observed anymore, the component(s) whose behavior were adapted are considered as (necessary) causes\footnote{Note that \cite{goessler2013generaltrace} distinguish between necessary and sufficient causality. In this paper however, we only focus on the former form, because the other algorithms do not make this distinction.}.
\par
In our implementation, we tried to stick to the original algorithm, but made two mentionable adaptations. One aspect, which is not addressed in the paper, is how to start the causality analysis. That is, the main question: Which sets of components should be analyzed for causality and in which order? The authors only specify how their proposed analysis is conducted with a given set $\mathcal{I}$, which indexes the components analyzed for being a cause \emph{together}. According to \cite{goessler2013generaltrace}, their algorithm makes the assumption that unwanted behavior in a system can only occur if at least one component does not behave as specified. Intuitively, it then makes sense to investigate the components, which did somehow violate their specification. Therefore, our implementation analyzes \emph{any} combination of components, which behaved incorrectly. That way, we do not miss out analyzing potential cause consisting of one ore more incorrect events. Moreover, we have implemented the algorithm to output sets of causing events to conform with our understanding of a cause (\ref{def:root_cause_mathematical}). Since \cite{goessler2013generaltrace} actually return sets of components, there is not a lot of change required. In our implementation, we determine the sets of causing components based on exactly the same algorithm proposed by \cite{goessler2013generaltrace}. Then, we extract the first violating event of each causing component. On that way, we transform each set of components into a set of events, each of which represents a hypothesized cause. In our opinion this seems natural, because \cite{goessler2013generaltrace} base their analysis on the first violation of a component, but don not explicitly return those events. A second adaptation is that we restricted the alphabet of this algorithm to strings. This reduces complexity and makes the implementation more convenient. In general however, a language over anything could be used by this algorithm.

\paragraph{Gössler and Astefanoaei \cite{goessler2014blaming}}
The concepts of this algorithm  are  similar to \cite{goessler2013generaltrace}. However, the main difference is that \cite{goessler2014blaming}  base their specification of the behavior of a system on \textit{timed automata}. Each component is represented as a timed automaton. During the causality analysis, potential incorrect behavior of one or more components is again replaced with behavior taken from their specification, in order to check, whether or not the occurred unwanted behavior can still be observed. The authors use \textit{Uppaal}\footnote{http://www.uppaal.org [Accessed 07 February 2017]} for the creation, graphical representation and analysis of timed automata.
\par
For this approach, no adaptations were needed. However, similarly as for the previous algorithm, \cite{goessler2014blaming} do not specify which components exactly to analyze for causality and the returned causes are again sets of components. For the solution of this problem, we have used the same approach as for \cite{goessler2013generaltrace} described in the above.

\paragraph{Mian et al.  \cite{mian2015auditing}}
Compared to the previous algorithms, this approach differs in two main aspects. First of all, it is not a causality analysis approach, but rather a policy compliance framework. Secondly, \cite{mian2015auditing} have developed a framework supporting auditors and not a pure algorithm. Although the concepts of a causality and policy compliance approach differ to some extent, we think a comparison of both is still reasonable in order to find exactly those differences in terms of effectiveness and efficiency. However, due to the fact that \cite{mian2015auditing} describes rather a framework, some adaptations were necessary in order to ensure comparability with the other algorithms. Specifically, we have developed an approach, which can check whether or not specific predefined rules of a system were met. An example for such a rule could be that a door is only allowed to be unlocked by a key card, if the holder of the key card is authorized to enter. All those events violating a rule are finally reported by the algorithm and interpreted as causes. As a result, our implementation is considerably different from the original auditing framework, which is why we will from now on refer to the latter as policy compliance algorithm inspired by concepts of \cite{mian2015auditing}.

\section{Case Study}\label{section:comparison}
In this section, we evaluate the implemented algorithms in a case study using \textit{ACCBench}. Thereby, an intelligent Door Control System will serve as an example. Firstly, an introduction to  the door system will be provided (Section \ref{subsection:introduction}). Then, we describe the used event logs (Section \ref{subsection:preparation}). Lastly, we compare and evaluate the results of executing the algorithms.

\subsection{Introduction}\label{subsection:introduction}
Let us consider our exemplary Door Control System closely. We have chosen this system for two main reasons. Firstly, it consists of multiple components with different specifications, therefore, offers a variety of different (failure) scenarios. Secondly, such a door system is a real-life socio-technical system. It demonstrates that accountability and causality are valuable properties of modern systems.
\par
The system used in this paper relies on specifications\footnote{\url{http://www.ucl.ac.uk/estates/security/specifications/Gallagher-System-Specification_v1.pdf} [Accessed 18 October 2015], \url{http://www.ucl.ac.uk/estates/maintenance/fire/documents/UCLFire_TN_001.pdf} [Accessed 05 September 2016]} by the University College London (UCL) and documents\footnote{\url{https://security.gallagher.com/gallagher-downloads/get/<id>},  where <id> $\in$ \{41, 44, 45, 57, 126\} [Accessed 05 September 2016]} from \textit{Gallagher Security}, which is the manufacturer of most of the described system-components.
\paragraph{Structure}
Intuitively, the functionality of a Door Control System is clear: It shall prevent unauthorized access within a building. For example, it might be a requirement that certain areas should only be accessible by a specific group of users. Hence, several technical components are needed to achieve these goals. 

\begin{wrapfigure}{R}{0.4\textwidth}
\vspace{-3mm}
\centering
\includegraphics[trim=6mm 6mm 6mm 6mm, clip, width=0.4\textwidth, keepaspectratio]{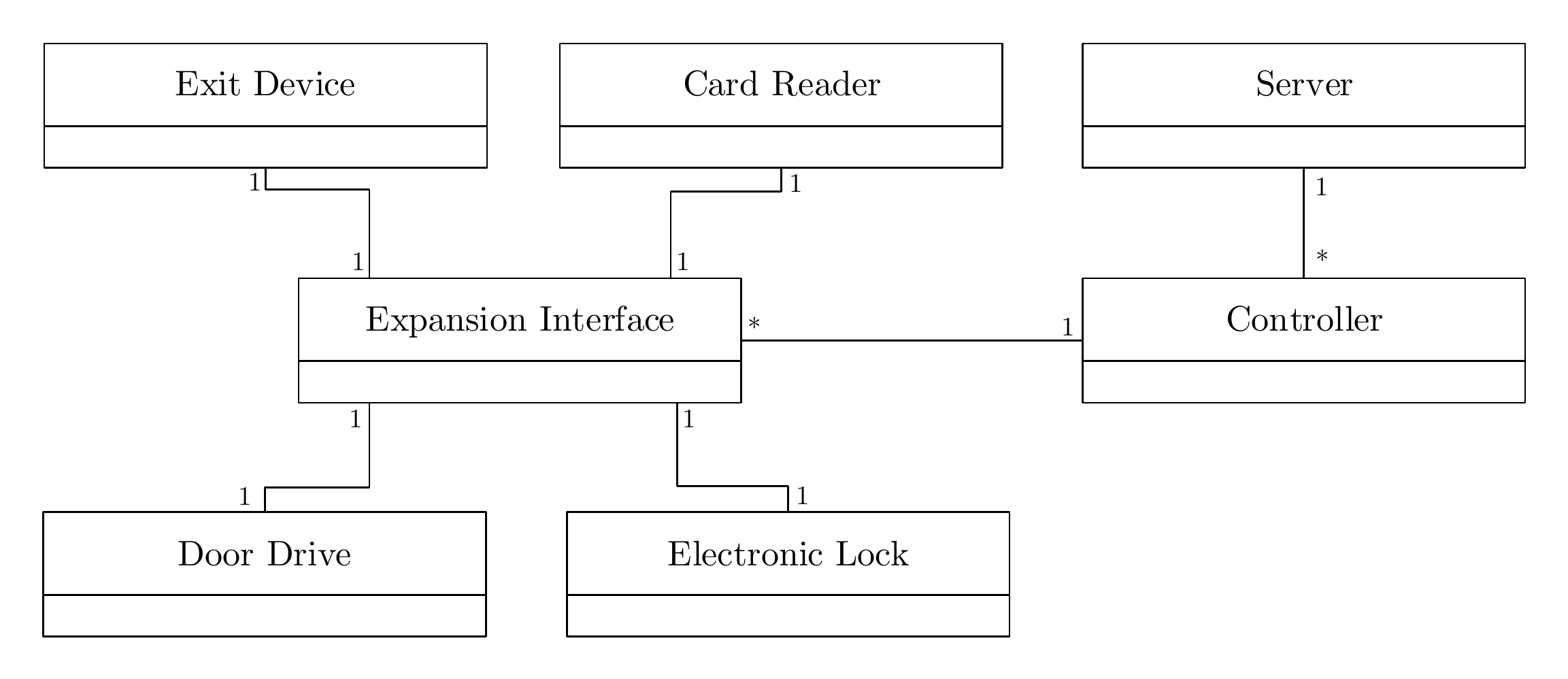}
\caption{Basic Door Configuration and Structure}
\label{fig:ucl_door_configuration}
\vspace{-4mm}
\end{wrapfigure}

\noindent
In the specification documents of UCL, a basic configuration of the doors is shown. We have modeled this configuration in Fig. \ref{fig:ucl_door_configuration}. Each door is fitted with a Card Reader (insecure side, outside) and an Exit Device (secure side, inside), which opens the door without having to use a key card. Furthermore, an Electronic Lock is attached to each door. We additionally extend this setting with a Door Drive, which automatically opens and closes a door. These four devices are connected to an Expansion Interface, i.e. a component, which allows to combine several other devices and forward their signals. Each Expansion interface communicates with a Controller connected to a central Server. Following the specifications of \textit{Gallagher Systems}, this Server is the main decision-making unit (e.g. for checking a user's a key card), yet also the connected Controllers have some of those capabilities in case they are currently not connected to the Server. However, for the sake of simplicity, we assume that the Controller is just a component receiving and forwarding instruction information from/to the Server to/from devices. Therefore, only the Server maintains the main logic of the door.

Each of the mentioned components writes a log entry when (1) an instruction has been received, (2) a received instruction has been forwarded, (3) a new instruction has been created and sent out, or (4) a (general) event occurred, e.g. a key card is detected to be invalid. Each log entry consists of the name of an event, the component it occurred at, a timestamp as well as an ID. As specified in Section \ref{section:accbench}, \textit{ACCBench} requires the event logs to be in the XES format.

In our system, we assume two basic situations of unwanted behavior (Def. \ref{def:unwanted_behavior}) whose cause(s) shall be determined. The first is \textit{Harm of Individuals} (E1). Obviously, such a system should be designed in a way so that it does not injure its users. However, it might happen that the door does not unlock and/or open. Especially in emergency situations like fire, this might block escape routes and therefore potentially harm people. Furthermore, individuals could get hurt, if the door closes unexpectedly, i.e. too quick after it has been opened. The second unwanted behavior is \textit{Unauthorized Access} (E2). For example, a person who should not be allowed in a restricted area, can access it. Reasons for this may be that the door is not closed at all or not locked. There are a variety of different causes, which may lead to specific unwanted behavior. For instance, a door might not be locked, because the Electronic Lock is damaged, but it could also be that the lock instruction was not sent to the latter.

\subsection{Preparation}\label{subsection:preparation}
To be able to compare the algorithms in the Door Control System, we need to (1) model its behavior as required by the algorithms (i.e. the configurations as shown in Fig. \ref{fig:accbench:informal_model}) and (2) obtain event logs containing wrong behavior.
\par
The configurations correspond to the requirements of each algorithm as explained earlier. That is, for the algorithm in \cite{goessler2013generaltrace} a language-based and for \cite{goessler2014blaming} a timed automata-based behavioral model of the system is provided. For our policy compliance approach based on concepts of \cite{mian2015auditing} we have specified a set of rules. 
\par
The event logs have been generated using \textit{CPNTools} and an approach similar to \cite{medeiros2005process}. We conducted a hazard and threat analysis using fault and attack trees to obtain relevant scenarios. In total, we generated 46 logs, of which 34 were created for analyzing the effectiveness and the remaining twelve for the performance. In the former case logs consist of between seven and 58 entries and contain at most a single complete interaction with the system. That is, at most one scenario starting with a user holding his key card against the reader and ending with the closing and locking of the door provided that user was authorized. Otherwise, the interactions stops with detecting that the user is not authorized. For the performance analysis, the number of log entries is between 550 and 2750. 

\begin{wrapfigure}{R}{0.35\textwidth}
\vspace{-3mm}
\centering
\includegraphics[trim=6mm 6mm 6mm 6mm, clip, width=0.35\textwidth, keepaspectratio]{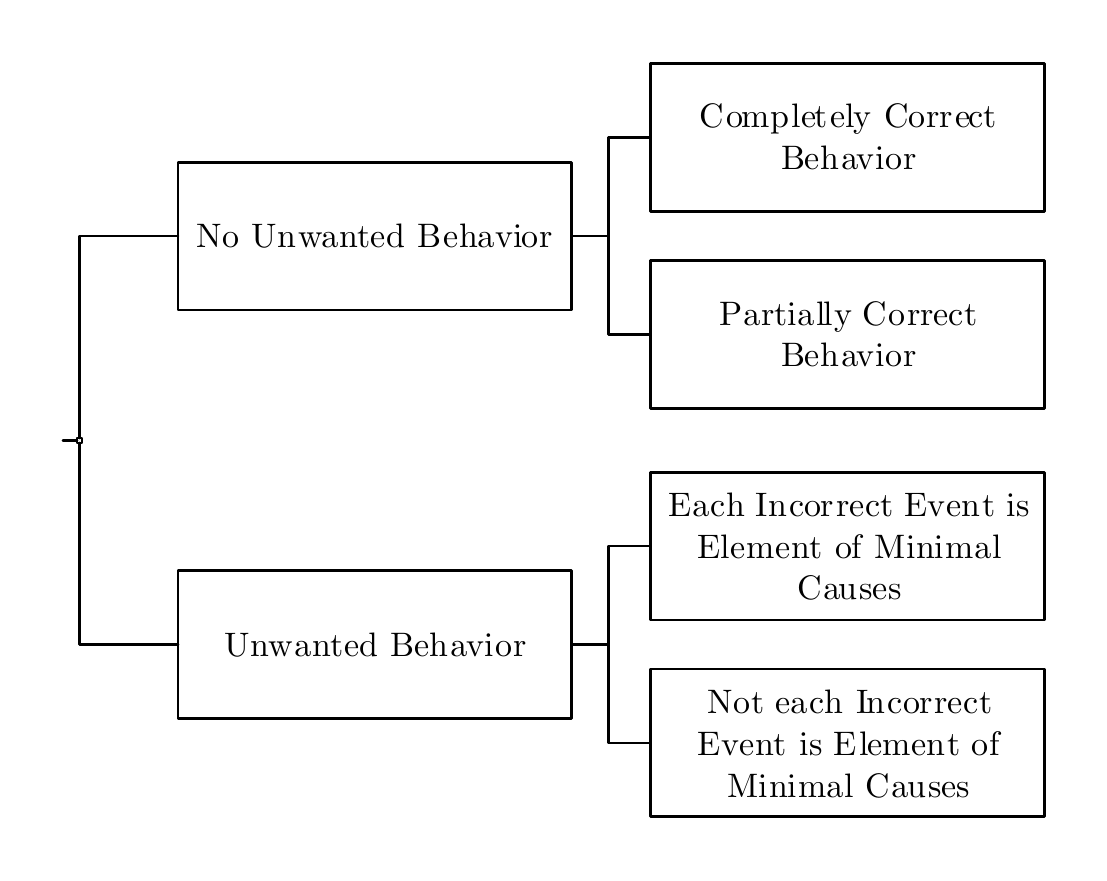}
\caption{Categorization of the Logged Scenarios}
\label{fig:taxonomy_behavior_dcs}
\vspace{-4mm}
\end{wrapfigure}

For a better interpretation of the results, we categorized the logged scenarios (Fig. \ref{fig:taxonomy_behavior_dcs}). We distinguish between scenarios, in which an unwanted behavior occurred, and scenarios, in which \emph{no} unwanted behavior occurred. However, in the latter case it is still possible that components did violate their specification, but this did not cause an unwanted behavior.

\noindent
Obviously, if no unwanted behavior occurred, even if some components behaved incorrectly, the causality algorithm should not report one or more potential causes. For a categorization in case unwanted behavior, we introduce the term \textit{minimal cause}, which we define as the smallest set of events causing respective unwanted behavior. 
Note that in the original theory of the two causality algorithms in \cite{goessler2013generaltrace} and \cite{goessler2014blaming} minimality of the returned causes is not addressed and as a consequence, we did not address it in our implementation. That is, the algorithms always return minimal and possibly non-minimal causes. Now, in our classification, we distinguish, whether or not each incorrect event\footnote{We refer to an event as incorrect, if it violates the specification of the component it occurred at.} in a log is element of a minimal cause for the current scenario. This allows us to see, if a causality algorithm can clearly detect that in some cases not necessarily each incorrect event alone or each combination of incorrect events can be blamed for causing unwanted behavior.

\subsection{Evaluation}
Having prepared the configurations for the three algorithms and generated the event logs, now we describe their evaluation and benchmarking. On the one hand, we will show the quantitative results obtained from \textit{ACCBench}, i.e. the effectiveness and performance, and on the other hand, we will also consider some more qualitative aspects of the algorithms.
\par
Before starting the evaluation, let us consider some important remarks. In particular when comparing the effectiveness of the algorithms, we need to take into account that causality algorithms might be quite different, especially concerning their individual definition of a cause. We have seen that the algorithms by \cite{goessler2013generaltrace} and \cite{goessler2014blaming} were designed to detect the cause(s) for observed unwanted behavior. In contrast, both the original approach of \cite{mian2015auditing} and our adapted implementation are only capable of detecting the violation of defined policies, yet the algorithm does not have a notion of unwanted behavior, which may result from a policy violation. The difference is important: Unwanted behavior results from violation(s) of policies, specifications, rules etc. but not each violation necessarily leads to unwanted behavior. We think that in general the comparison of causality algorithms regarding the correctness of results is only reasonable, if (1) one's own definition of a cause is congruent with the algorithms' definition and (2) all the algorithms share the same definition of a cause. 
Our implementations of the algorithms in \cite{goessler2013generaltrace} and \cite{goessler2014blaming} share the same definition of a cause and therefore a comparison of their reported results and the  expected results is reasonable. However, this is not the case for our implementation of the policy compliance approach in \cite{mian2015auditing}. Nonetheless, we think that it makes sense to compare the results of this algorithm with our expected results to understand the difference between a policy compliance algorithm and an actual causality algorithm. Furthermore, this creates a baseline for a causality algorithm: The latter should be more effective than a policy compliance algorithm.
\par
Moreover, we do not claim that the results of this case study generalize. The results only show a specific application of the algorithms and therefore our findings may be different for other systems. Nevertheless, we think that our example can to some extent show strengths and weaknesses of the analyzed algorithms and point out areas for future improvements.

\paragraph{Effectiveness}
As described in Section \ref{subsection:metrics}, two approaches for estimating the effectiveness of an algorithm are integrated into \textit{ACCBench}. We will present results for both. We found out that comparing the effectiveness of the algorithms on category-level of our categorization (Fig. \ref{fig:taxonomy_behavior_dcs}) shows some insights. Hence, we conducted our analysis separately for each category.
\begin{wrapfigure}{R}{0.45\textwidth}
\vspace{-3mm}
\centering
\begin{tikzpicture}
\begin{axis}[counter_barchart_y]
\addplot coordinates {(Gößler and Astefanoaei, 1) (Gößler and Métayer, 1) (Policy Compliance Algorithm, 0.3333)}; 
\addplot coordinates {(Gößler and Astefanoaei, 1) (Gößler and Métayer, 1) (Policy Compliance Algorithm, 1.0)}; 
\addplot coordinates {(Gößler and Astefanoaei, 1) (Gößler and Métayer, 1) (Policy Compliance Algorithm, 0.3333)}; 

\addplot coordinates {(Gößler and Astefanoaei, 1) (Gößler and Métayer, 1) (Policy Compliance Algorithm, 0.3333)}; 
\addplot coordinates {(Gößler and Astefanoaei, 1) (Gößler and Métayer, 1) (Policy Compliance Algorithm, 1.0)}; 
\addplot coordinates {(Gößler and Astefanoaei, 1) (Gößler and Métayer, 1) (Policy Compliance Algorithm, 0.3333)}; 
\legend{$F_1$, Recall,Precison, $F_1$', Recall',Precison'}
\end{axis}
\end{tikzpicture}
\caption{Metrics for the categories \textit{Completely Correct Behavior} and \textit{Partially Correct Behavior}}
\label{fig:metrics_completely_partially}
\vspace{-4mm}
\end{wrapfigure}
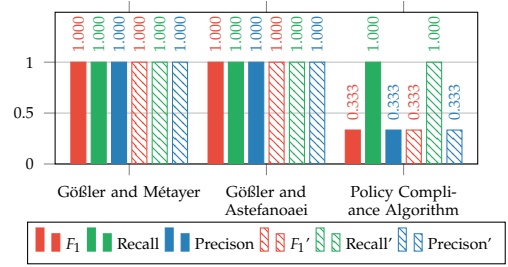

\begin{wrapfigure}{R}{0.45\textwidth}
\vspace{-3mm}
\centering
\begin{tikzpicture}
\begin{axis}[counter_barchart_y]
\addplot coordinates {(Gößler and Astefanoaei,0.8889) (Gößler and Métayer, 0.9556) (Policy Compliance Algorithm, 0.8667)}; 
\addplot coordinates {(Gößler and Astefanoaei,0.9333) (Gößler and Métayer, 1) (Policy Compliance Algorithm, 0.8667)}; 
\addplot coordinates {(Gößler and Astefanoaei,0.9333) (Gößler and Métayer, 0.9333) (Policy Compliance Algorithm, 0.8667)}; 

\addplot coordinates {(Gößler and Astefanoaei,0.8889) (Gößler and Métayer, 0.9556) (Policy Compliance Algorithm, 0.9111)}; 
\addplot coordinates {(Gößler and Astefanoaei,0.9333) (Gößler and Métayer, 1) (Policy Compliance Algorithm, 0.9333)}; 
\addplot coordinates {(Gößler and Astefanoaei,0.9333) (Gößler and Métayer, 0.9333) (Policy Compliance Algorithm, 0.9)}; 
\legend{$F_1$, Recall,Precison, $F_1$', Recall',Precison'}
\end{axis}
\end{tikzpicture}
\caption{Metrics for category \textit{Each Violating Event is an Element of Minimal Cause}}
\label{fig:metrics_each_violating}
\vspace{-4mm}
\end{wrapfigure}
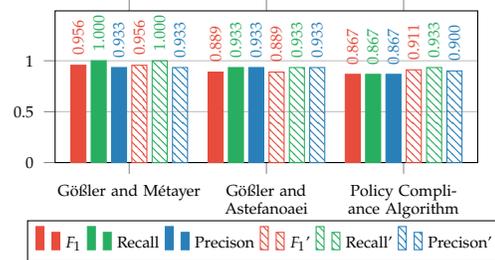

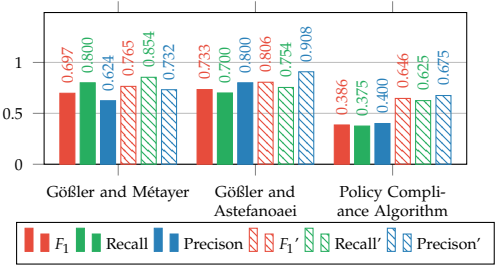
\begin{wrapfigure}{R}{0.45\textwidth}
\vspace{-3mm}
\centering
\begin{tikzpicture}
\begin{axis}[counter_barchart_y]
\addplot coordinates {(Gößler and Astefanoaei,0.7333) (Gößler and Métayer, 0.6972) (Policy Compliance Algorithm, 0.3857)}; 
\addplot coordinates {(Gößler and Astefanoaei,0.7) (Gößler and Métayer, 0.8) (Policy Compliance Algorithm, 0.375)}; 
\addplot coordinates {(Gößler and Astefanoaei,0.8) (Gößler and Métayer, 0.6238) (Policy Compliance Algorithm, 0.4)}; 

\addplot coordinates {(Gößler and Astefanoaei,0.8055) (Gößler and Métayer, 0.7649) (Policy Compliance Algorithm, 0.6464)}; 
\addplot coordinates {(Gößler and Astefanoaei,0.7541) (Gößler and Métayer, 0.8541) (Policy Compliance Algorithm, 0.625)}; 
\addplot coordinates {(Gößler and Astefanoaei,0.9083) (Gößler and Métayer, 0.7321) (Policy Compliance Algorithm, 0.675)}; 
\legend{$F_1$, Recall,Precison, $F_1$', Recall',Precison'}
\end{axis}
\end{tikzpicture}
\caption{Metrics for category \textit{Not Each Violating Event is an Element of Minimal Cause}}
\label{fig:metrics_not_each_violating}
\vspace{-4mm}
\end{wrapfigure}

We start with the event logs, in which no unwanted behavior can be observed, i.e. we combine both the fully and partially correct logs. As seen in Fig. \ref{fig:metrics_completely_partially}\footnote{The metrics marked with (') are based on the modified binary classification. This also applies to Figs. \ref{fig:metrics_each_violating} \& \ref{fig:metrics_not_each_violating}.}), there is a considerable difference between causality algorithms and the policy compliance algorithm. As explained before, the latter approach has no notion of whether or not unwanted behavior actually occurred and therefore blames any incorrect event. This decreases the \textit{precision} metrics and thus the $F_1$-measure. Notice that the average \textit{recall} is equal to 1 in this category, because the number of false negatives will always be zero due to the mere fact that the set of expected causes is empty. The reason why the other two metrics are not equal to zero for our policy compliance algorithm is that in the completely correct scenarios the latter does not report any violation and therefore matches the expected result. Since the other two algorithms can always detect that no unwanted behavior occurred in all the scenarios of this category, their results are fully correct.
\par
For event logs in which unwanted behavior can be observed, we start with the analysis of those where each incorrect event is an element of a minimal cause. Computing again our metrics, this results in the values shown in Fig. \ref{fig:metrics_each_violating}. 

\noindent
We can see that our implementations of the two causality algorithms had some problems in always detecting a fully correct result, which is why the corresponding metrics do not evaluate to 1 anymore. Our analysis has shown that there exist multiple reasons for these results. The language-based causality algorithm \cite{goessler2013generaltrace} seems to have the problem of replacing the incorrect behavior of components in the counterfactual scenario, although this behavior should not change. As a result, the faulty components and their events might be blamed just because during their analysis wrong behavior of other components is removed. For our implementation of the algorithm in \cite{goessler2014blaming}, we observed the same problem. Furthermore, we detected that this approach is not fully correct and can produce different results in some situation than it claims to. This issue can arise if broadcast channels are used in the timed automata modeling a system's behavior. Additionally, we detected that deadlocks in the networks of timed automata constructed by the algorithm of \cite{goessler2014blaming} prevent the proper construction of counterfactual behavior and lead to wrong hypothesized causes. We have verified the existence of this problem during a discussion with the first author of \cite{goessler2014blaming}. Lastly, we can see that our policy compliance approach performs almost equally as compared to the previous algorithms. However, finding different variants of scenarios in the current category was difficult, which is why in most of them only one component is behaving incorrectly with a single event. Thus, the policy compliance algorithm reports mostly a correct result, even though it is not based on the same definition of a cause.

Finally, let us consider those event logs, in which not each violation is an element of a minimal cause. Figure \ref{fig:metrics_not_each_violating} shows again that there are differences between all three algorithms. We can see that the timed automata-based causality algorithm yields slightly better results than the language-based one, but both algorithms perform worse than in previous categories. The effectiveness metrics for our policy compliance approach decreased significantly as well. All the problems described in the above also occurred in this class of logs, but, due to the diversity of the scenarios and corresponding logs, more often. Moreover, we found that the two causality algorithms ignore other incorrect behavior of a component once it has violated its specification. That is, even if there has been incorrect behavior, which did not lead to unwanted system behavior, any potentially wrong behavior coming afterwards is not considered. Since the policy compliance approach simply returns any violation of a rule as sets of single causes, yet the event logs in the current category expose more sophisticated causes for the unwanted behavior.

\vspace{-2mm}
\paragraph{Performance}
To obtain insights on the performance in terms of execution time and memory allocation, we analyzed on the one hand the 34 event logs used for evaluating the effectiveness as well as twelve logs consisting of a higher amount of events (see Section \ref{subsection:preparation}). In three of the latter logs, no unwanted behavior can be observed. Basically, we just looped one interaction with the Door Control System, i.e. from using a key card until the door is closed again, multiple times. The remaining nine event logs contain unwanted behavior caused by a single event and differ in their length and the position of this single incorrect event, i.e. in the beginning, middle or end of the log.
\par
All our measurements were conducted with \textit{ACCBench} on a machine equipped with an Intel\textregistered~Core\texttrademark~i7-4700HQ CPU, 8GB RAM and running Windows 10. For space reasons, we only show our results concerning the execution time of the algorithms. 
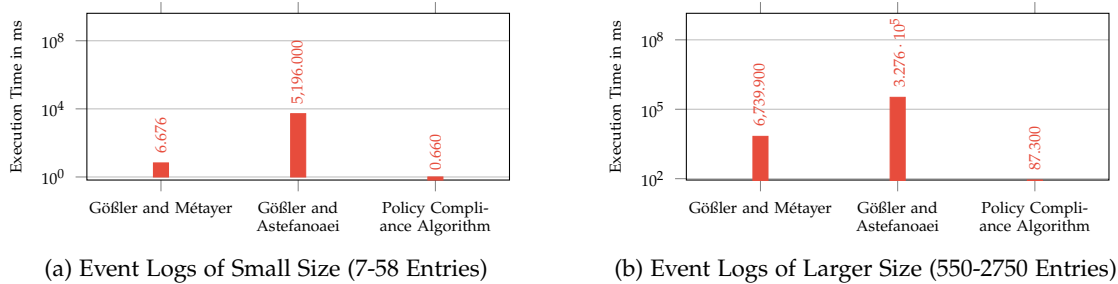
\begin{figure}[ht!]
\vspace{-3mm}
\centering
\subfloat[Event Logs of Small Size (7-58 Entries)]{%
\begin{tikzpicture}
\begin{axis}[counter_barchart_y, width=0.45\textwidth, ymode=log, ymax=52000, point meta=rawy, y tick label style={/pgf/number format/1000 sep=, font=\tiny}, height=3.8cm, ylabel=Execution Time in ms, label style={font=\tiny}, enlarge y limits={1,upper},]
\addplot coordinates {
(Gößler and Astefanoaei, 5196) 
(Gößler and Métayer, 6.676) 
(Policy Compliance Algorithm, 0.6598)
};
\end{axis}
\end{tikzpicture}
}
\hspace{0.05\textwidth}
\subfloat[Event Logs of Larger Size (550-2750 Entries)]{%
\begin{tikzpicture}
\begin{axis}[counter_barchart_y, width=0.45\textwidth, ymode=log, ymax=350000, point meta=rawy, y tick label style={/pgf/number format/1000 sep=, font=\tiny}, height=3.8 cm, ylabel=Execution Time in ms, label style={font=\tiny}, enlarge y limits={1,upper}]
\addplot coordinates {
(Gößler and Astefanoaei, 327641.2) 
(Gößler and Métayer, 6739.9) 
(Policy Compliance Algorithm, 87.3)
};
\end{axis}
\end{tikzpicture}
}
\vspace*{-2mm}
\caption{Average Execution Times for the Analysis of a Single Event Log}
\label{fig:performance}
\vspace{-4mm}
\end{figure}
As seen in (Fig. \ref{fig:performance}), the size of log in terms of the number of events it contains, increases the execution time for each algorithm. However, whereas the execution time for our implementation of the language-based algorithm \cite{goessler2013generaltrace} and our policy compliance approach stays in the range of seconds or milliseconds, the timed automata-based algorithm \cite{goessler2014blaming} exceeds five minutes on average for larger logs. The reason seems to be the usage of \textit{Uppaal} and the model checking on timed automata. We also found out that it plays a role for the two causality algorithms in which part of a log incorrect behavior occurs, which potentially leads to unwanted system behavior. The reason is that both of these algorithms try to generate their counterfactual scenarios while keeping a specific prefix of the actually observed behavior in the event log. As a result, more data needs to be considered during the computation of counterfactual system behavior. Furthermore, the number of misbehaving components plays a role for the execution time as well. Since all combinations of components are analyzed to obtain which possibly caused unwanted behavior, the number of analysis iterations of the applied algorithm increases exponentially. Interestingly, all the previous factors do not apply to our policy compliance approach. It analyzes each event of a log once no matter how many components violated their specification. As we can see, this yields a higher performance in terms of execution time as compared to the actual causality algorithms by \cite{goessler2013generaltrace} and \cite{goessler2014blaming}.

\vspace{-2mm}
\paragraph{Qualitative Analysis}
For our qualitative analysis, we decided on three dimensions: The effort for creating the configuration for the algorithm, i.e. the modeling of the system behavior, limitations an algorithm might have and potential dependencies. Our results are summarized in Tab. \ref{tab:qualitative_analysis}.

\begin{table}[ht!]
\centering
\footnotesize
\begin{tabular}{p{4.9cm} | p{4.9cm} | p{4.9cm}}
\multicolumn{1}{c |}{\textbf{Gößler and Métayer \cite{goessler2013generaltrace}}} &\multicolumn{1}{c |}{\textbf{Gößler and Astefanoaei \cite{goessler2014blaming}}} &\multicolumn{1}{C{4.22cm}}{\textbf{Policy Compliance Algorithm}} \\ \hline
\vspace{-3mm}\begin{itemize}[leftmargin=2.5mm]
\setlength{\itemsep}{-0.3em}
\renewcommand\labelitemi{--}
\item Configuration Effort: Language-based configuration is rather abstract, complex and error-prone. Possibly good idea, if formal specification of a system exists.
\item Limitations: Specifically for our implementation: language only over strings (components) and tuples of strings (complete system), i.e. time is not considered, only the order of events (In general: Language over anything theoretically possible). Allowed behavior has fixed starting point.
\item Dependencies: None
\end{itemize}

&\vspace{-4mm}\begin{itemize}[leftmargin=2.5mm]
\setlength{\itemsep}{-0.3em}
\renewcommand\labelitemi{--}
\item Configuration Effort: Modeling with timed automata rather intuitive and easy validation, e.g. through simulation. May become complex with increasing number of components and/or complex behavior.
\item Limitations: Makes the assumption that interaction between components is synchronous. Specifications of component have fixed starting point. Only event and its timestamp can be considered during the analysis (not relevant for this case study).
\item Dependencies: \textit{Uppaal}; API problematic under Linux.
\end{itemize}

&\vspace{-4mm}\begin{itemize}[leftmargin=2.5mm]
\setlength{\itemsep}{-0.3em}
\renewcommand\labelitemi{--}
\item Configuration Effort: Intuitive, if few events. May become complex for many rules and/or complex constraints.
\item Limitations: Only event and its timestamp can be considered during the analysis (not relevant for this case study). Complex rules potentially not possible.
\item Dependencies: None
\end{itemize}
\\ \hline
\end{tabular}
\caption{Summary of the Qualitative Analysis}\label{tab:qualitative_analysis} 
\vspace*{-5mm}
\end{table}
\section{Conclusion and Future Work}\label{section:conclusion}
In this paper, we proposed \textit{ACCBench}, a novel and extensible benchmark tool allowing to compare causality algorithms. This comparison applies metrics concerning the effectiveness and the performance of an algorithm. For the effectiveness, we developed two criteria based on binary classification, which try to estimate how effectively an algorithm can find the causing event(s) for unwanted behavior (Def. \ref{def:unwanted_behavior}). Additionally, we implemented two recent causality algorithms described in \cite{goessler2013generaltrace} and \cite{goessler2014blaming} as well as a policy compliance approach in which we integrated ideas and concepts of \cite{mian2015auditing}. Finally, we evaluated these algorithms using \textit{ACCBench} in a case study. 
\par
We concluded that the two causality algorithms report better results, yet in some cases with heavy performance impact compared to the policy compliance approach. Moreover, we found that the type of input and information a user needs to provide to an algorithm have a significant impact on the applicability. For example, we noted that, the language-based framework of \cite{goessler2013generaltrace} for modeling a system is powerful but quite abstract approach, whereas the usage of timed automata in \cite{goessler2014blaming} might be more intuitive. The rules in our policy compliance algorithm seemed promising as well, but might become complex and less applicable for other systems.
\par
We believe that this paper provides an interesting and relevant contribution to accountability and causality research. In the future, we would like to extend our work and address several additional aspects. Because our case study was a simplified example using simulated log files, we have only shown the applicability of the algorithms in a rather isolated setting. Hence, we would like to apply the algorithms in an existing system  to see if and how much pre-processing is required. In general, more research is required to achieve meaningful statements about the quality of the algorithms considered in this paper. The case study has also shown that the performance of the algorithms decreases the larger the event logs get. Thus, we plan to investigate approaches for addressing performance issues.

\section*{Acknowledgment}\label{section:ack}
\vspace*{-4mm}
This work is part of the TUM Living Lab Connected Mobility (TUM LLCM) project and has been funded by the Bavarian Ministry of Economic Affairs and Media, Energy and Technology (StMWi) through the Center Digitisation.Bavaria, an initiative of the Bavarian State Government.

\bibliographystyle{eptcs}
\bibliography{bibliography_paper}

\end{document}